\documentclass[fleqn,10pt]{wlscirep}
\usepackage[utf8]{inputenc}
\usepackage[T1]{fontenc}
\usepackage{caption}
\usepackage{subcaption}
\usepackage{amsmath}
\usepackage{textcomp}
\usepackage{comment}
\usepackage{graphicx}
\usepackage{booktabs} 

\newcommand{\refframe}[1]{#1}
\renewcommand{\vec}[1]{\boldsymbol{#1}}
\newcommand{\unitvec}[1]{\hat{\vec{#1}}}
\newcommand{\mat}[1]{\mathbf{#1}}
\newcommand{\tpose}{\mathsf{T}}
\newcommand{\vecframe}[2]{\vec{#1}^{\refframe{#2}}}
\newcommand{\rotation}[2]{\mat{R}_{\refframe{#1}}^{\refframe{#2}}}

\title{End-to-end example-based sim-to-real RL policy transfer based on neural stylisation with application to robotic cutting}
\author[1,2,$\dagger$]{Jamie Hathaway}
\author[1,3,$\dagger$,*]{Alireza Rastegarpanah}
\author[1,2]{Rustam Stolkin}
\affil[1]{Extreme Robotics Lab, School of Metallurgy and Materials, University of Birmingham, Birmingham, B15 2TT, UK}
\affil[2]{The Faraday Institution, Quad One, Harwell Science and Innovation Campus, Didcot, OX11 0RA, UK}
\affil[3]{School of Computer Science and Digital Technologies, Aston University, Birmingham B4 7ET, UK}
\affil[*]{Corresponding Author: Alireza Rastegarpanah, Email: a.rastegarpanah@bham.ac.uk}
\affil[$\dagger$]{Jamie Hathaway and Alireza Rastegarpanah are both recognised as joint first authors.}

 
\begin{abstract}
Whereas reinforcement learning has been applied with success to a range of robotic control problems in complex, uncertain environments, reliance on extensive data - typically sourced from simulation environments - limits real-world deployment due to the domain gap between simulated and physical systems, coupled with limited real-world sample availability. We propose a novel method for sim-to-real transfer of reinforcement learning policies, based on a reinterpretation of neural style transfer from image processing to synthesise novel training data from unpaired unlabelled real world datasets. We employ a variational autoencoder to jointly learn self-supervised feature representations for style transfer and generate weakly paired source-target trajectories to improve physical realism of synthesised trajectories. We demonstrate the application of our approach based on the case study of robot cutting of unknown materials. Compared to baseline methods, including our previous work, CycleGAN, and conditional variational autoencoder-based time series translation, our approach achieves improved task completion time and behavioural stability with minimal real-world data. Our framework demonstrates robustness to geometric and material variation, and highlights the feasibility of policy adaptation in challenging contact-rich tasks where real-world reward information is unavailable.
\end{abstract}
\begin{document}

\flushbottom
\maketitle
\thispagestyle{empty}

\section*{Introduction}
Emerging applications for robotics have fostered increasing interest in low-volume, high-mix disassembly processes in industry. These processes are characterised by a high degree of uncertainty coupled with demands of logistical flexibility, which traditionally implies the requirement for expensive reprogramming and reconfiguration of robots. This is of interest in domains such as nuclear decommissioning, robotic disassembly of complex products for recycling and re-use, and even areas such as robotic surgery or demolition with robotised demolition equipment. Nonetheless, challenges exist in automated planning and task execution for destructive operations. Whereas manufacturing paradigms centre around achieving high dimensional tolerances and precise control on a known workpiece, for disassembly, the precise location of cutting is less important (few mm as opposed to $\mu$m) while the precise sequence of cutting operations may not be known in advance. This uncertainty has motivated various approaches to robotic cutting, consisting of goal-conditioned trial-and-error \& revision \cite{CognitiveRoboticsBasic, CognitiveRoboticsRevision}, 3D reconstruction \& planning \cite{Cutting:RoboticMilling3DPointCloud}, and online learning \& adaptation \cite{ContourErrorCNCLearningRL, Cutting:FeedRateCharacterisation}.

Reinforcement learning (RL) has been applied with success to a variety of contact-rich tasks \cite{ComplianceControlPegInHoleRL, VariableImpedanceActionSpaceRL}, including robotic cutting \cite{RLMachiningDeformationControl, LearningRoboticMillingRL}, particularly with difficult-to-model environments with complex robot-environment interactions, but are nonetheless data intensive. Whereas simulation environments offer reduced complexity and overhead of data collection, differences between simulated and physical cutting processes limit the applicability of adaptive methods to real-world tasks. Examples of such differences include motor backlash, tool wear, chattering, cross-domain mismatch of process and model parameters and other disturbances. These differences motivate the use of domain adaptation methods to align representations or behaviours across domains with minimal real-world supervision. These can be broadly separated into unified feature representation learning, model-based correction, and model-free synthesis of target domain examples. 

Domain adaptive methods include \cite{DomainAdaptationRLUnifiedLatentRepresentation} in which policies are trained on a cross-domain latent feature representation by aligning source and target domain distributions. A related concept applied to milling was proposed in \cite{RLMachiningDeformationControl} based on a cross-domain meta-model, trained on pairwise unified feature representations. Similarly, adversarial losses using domain discriminators have been employed for cross-domain tool wear classification \cite{AdversarialDomainAdaptation}. Reconstruction-based methods have also been employed to jointly model observation and class distributions\cite{ReconstructionPsuedoCVAE}; this concept has been further developed based on conditional variational autoencoders (CVAEs) \cite{ModelBasedRLCVAE} wherein CVAE feature representations were used to train an RL policy, while feature representations are aligned across domains.

Model-based approaches have previously also been employed for domain adaptation, wherein a source domain task model is augmented with a corrective model based on physics-informed approaches \cite{IdentificationDisturbanceObserver}, neural networks \cite{SimToRealNeuralAugmentedSimulation, DeepInverseDynamicModelSim2Real} or Gaussian process (GP) models \cite{GPDisturbanceObserver, SimToRealBasedOnGPandDR} learned from target domain data. In our previous work, \cite{Cutting:SimToRealAdaptation} we proposed an imitation learning framework in which a GP corrective model was learned from multiple cutting demonstrations. Nonetheless, model-based approaches incur limitations of modelling assumptions under which the models are introduced, and incur a dataset overhead, particularly for deep predictive modelling approaches.

Relating to the aforementioned approaches is direct alignment of observations across domains via translation or generative models. In the context of milling, \cite{GenerativeNNBasedDomainAdaptationIncompleteTargetDomain} proposed a domain adaptation method for condition monitoring of different milling tools based on a generative CNN. Similarly, \cite{OneShotDomainAdaptiveImitationLearning}, proposed a domain adaptive imitation learning framework from visual demonstrations based on CycleGAN \cite{CycleGAN2017}. Generation at object level has also been proposed \cite{InstaceStyleTransfer6DPose} wherein a StyleGAN image translation model is trained object-wise on weakly-paired cross-domain datasets for 6D pose estimation. CVAEs have also been employed for domain adaptation via synthesis of novel target domain examples \cite{ZeroShotDomainAdaptationCVAE}. 

Neural style transfer has been extensively researched in the context of image processing \cite{NeuralStyleTransferOriginal, ControllingPerceptualStyleTransfer}. Recently, this concept has been extended to motion execution. Thus far, its application has been limited largely to expressive stylised motions mirroring that of human operators \cite{NeuralPolicyStyleTransfer, NeuralPolicyStyleTransferRL}. Nonetheless, its applicability to synthesise novel trajectories with characteristics of diverse human operators presents a compelling case for its application to other domain adaptation problems. Recently, this has been applied for dataset augmentation tasks \cite{StyleTime}. A limitation of the aforementioned methods is lack of a suitable pairing mechanism for style and content, as well as lack of feature extractor backbones prevalent in image processing tasks. For transfer learning, addressed this problem \cite{StyleTransferTransferLearning} by building on the concept of conditional adversarial domain adaptation \cite{DomainAdaptationConditionalAdversarial} to achieve feature-level style transfer for transfer learning. Nonetheless, adversarial alignment can be difficult to train, with well-known problems of mode collapse and vanishing gradients. Whereas these developments have been applied to time series classification problems, application of style transfer for RL policy transfer is, to the best of our knowledge, largely unexplored.

\begin{figure}[!t]
    \centering
    \includegraphics[width=0.85\textwidth]{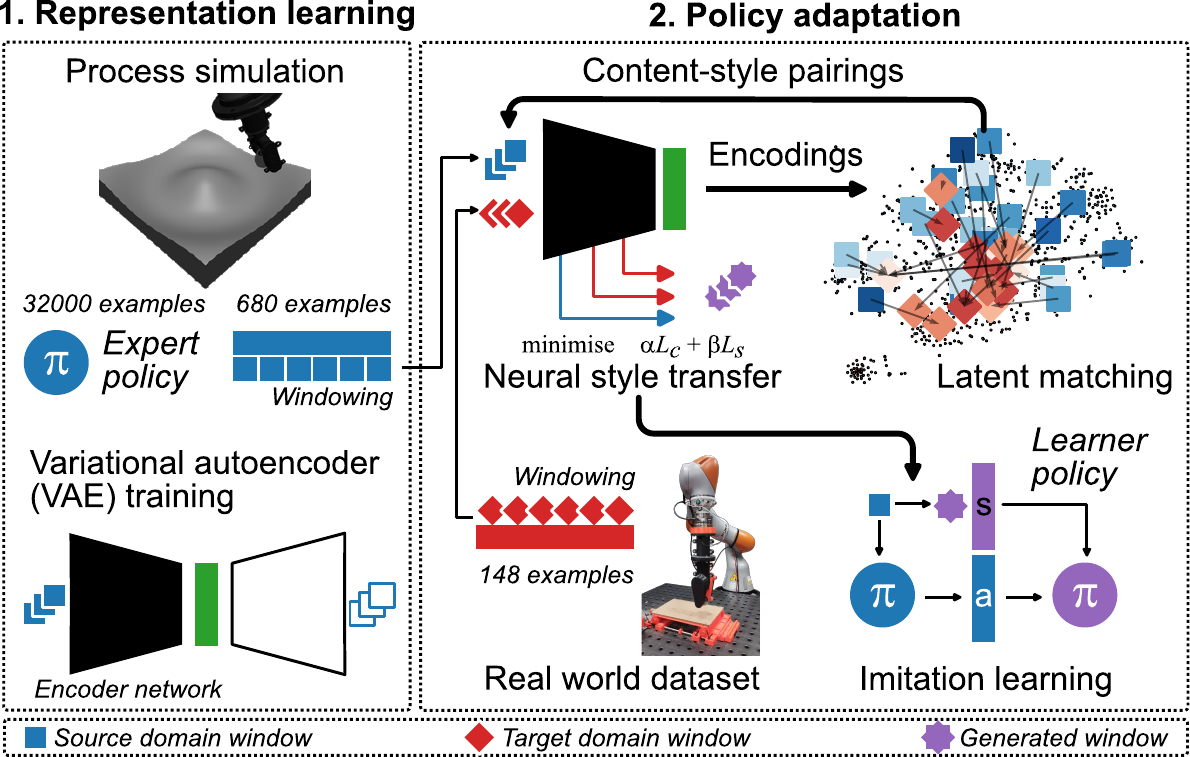}
    \caption{Overview of proposed framework. In the first stage, a simulation of cutting mechanics is used to generate an expert policy and a variational autoencoder (VAE) is trained on simulated trajectory windows. In the second stage, the VAE encoded representations are used to generate pairings between a simulated and real world dataset which are used as style targets. Finally, expert trajectories are used to train a learner target domain policy with the generated observation windows.}
    \label{fig:Introduction-Framework-Overview}
\end{figure}

This paper extends our previous example-based approach for sim-to-real adaptation to arbitrary real world examples. As with our previous work, our approach does not require re-training of classifiers or encoder networks to adapt to new scenarios (different disturbance forces, differing sensor dynamics, etc.). In contrast to prior work that applies neural style transfer primarily for stylised motion synthesis or dataset augmentation, we apply it as a trajectory-level domain adaptation mechanism for robotic skill transfer. Our contributions are threefold: (1) a latent-space pairing mechanism for content and style that operates without paired examples or retraining; (2) a novel transfer framework based on neural style transfer that does not require labelled or reward-supervised data from the target domain; and (3) empirical evaluation on robotic cutting, a task where conventional reinforcement learning pipelines are difficult to apply due to the absence of reward signal in the real-world deployment environment. An overview of our framework is provided in Figure \ref{fig:Introduction-Framework-Overview}.

\section*{Style transfer framework}
\subsection*{Variational autoencoder}\label{subsec:Method-Autoencoder}
\begin{figure}[t]
    \centering
    \includegraphics[width=0.5\linewidth]{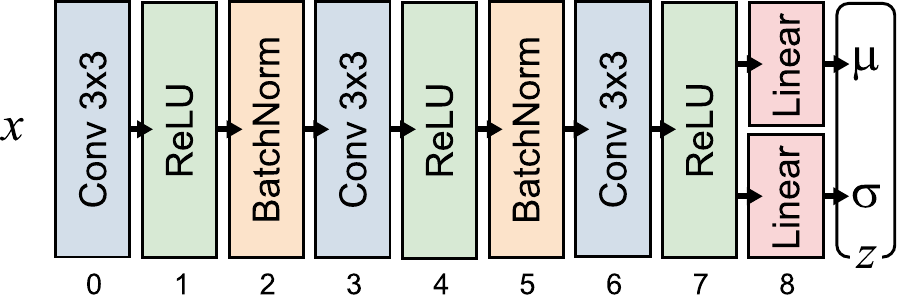}
    \caption{Overview of VAE encoder architecture; layer indices for style transfer are demarcated.}
    \label{fig:Method-VAE-Architecture}
\end{figure}

The variational autoencoder (VAE) consists of two neural networks: an encoder $q_\phi(\vec{z}|\vec{x})$ that approximates the posterior over latent variables $\vec{z} \in \mathbb{R}^L$, and a decoder $p_\theta(\vec{x}|\vec{z})$ that reconstructs the data from the latent representation. The encoder network $q_\phi(\vec{z}|\vec{x})$ outputs distributional parameters $\vec{\mu}_\phi(\vec{x})$ and diagonal log-variance $\log \vec{\sigma}^2_\phi(\vec{x})$ of a multivariate Gaussian posterior as
\begin{equation}
q_\phi(\vec{z}|\vec{x}) = \mathcal{N}(\vec{z}; \vec{\mu}_\phi(\vec{x}), \operatorname{diag}(\vec{\sigma}_\phi(\vec{x})))
\end{equation}
A latent code \( \vec{z} \) is sampled via the reparametrisation trick:
\begin{equation}
\vec{z} = \vec{\mu}_\phi(\vec{x}) + \vec{\sigma}_\phi(\vec{x}) \odot \vec{\epsilon}, \quad \vec{\epsilon} \sim \mathcal{N}(\vec{0}, \vec{I}).
\end{equation}
The decoder $p_\theta(\vec{x}|\vec{z})$ reconstructs the input $\vec{x}$ from the latent code; for continuous data, we used an isotropic Gaussian likelihood $\mathcal{N}(\vec{x}; \vec{\mu}_\theta(\vec{z}), \vec{I})$. The VAE loss function is expressed as the evidence lower bound (ELBO), which comprises a reconstruction loss and a KL divergence regularising term:
\begin{equation}
\mathcal{L}(\theta, \phi; \vec{x}) = \mathbb{E}_{q_\phi(\vec{z}|\vec{x})} [\log p_\theta(\vec{x}|\vec{z})] - D_{\mathrm{KL}}[q_\phi(\vec{z}|\vec{x}) \,\|\, p(\vec{z})]
\end{equation}
where $p(\vec{z}) = \mathcal{N}(\vec{0}, \vec{I})$ is the standard normal prior over latent codes. Training was carried out with the Adam optimiser, with model hyperparameters established via manual search, reported in Table \ref{tab:Method-Hyperparameters}. Both encoder and decoder networks were implemented as strided convolutional networks with 3 layers. Batch normalisation was further employed to accelerate convergence and reduce training instability. The encoder architecture is visualised in Figure \ref{fig:Method-VAE-Architecture}.

The VAE training dataset consisted of a mixture of 680 on-policy and off-policy simulated trajectories. We consider a trajectory as a multivariate time series of length $T$ which comprises a sequence of state-action pairs:
\begin{equation}
\tau = \{(\vec{x}_t, \vec{y}_t) \}_{t=1}^{T}
\end{equation}
where $\vec{x}_t \in \mathbb{R}^{N_{S}}$, $\vec{y}_t \in \mathbb{R}^{N_{A}}$ are the states, actions at time $t$ respectively. Each trajectory $\tau$ is divided into overlapping windows of length $N$, resulting in a set of state and action windows:
\begin{align}
x^{(i)}= & [x_{t},x_{t+1},\dots,x_{t+N}]\in\mathbb{R}^{N\times N_{S}}\\
y^{(i)}= & [y_{t},y_{t+1},\dots,y_{t+N}]\in\mathbb{R}^{N\times N_{A}}
\end{align}
The window width $N$ and latent code dimensionality emerge as tunable parameters for which a trade-off exists between the temporal context afforded to the model, reproduction accuracy and saliency of the latent space. Through preliminary experiments, this was reflected in increased RMS error of the autoencoder reconstructions and reduced average cosine similarity between simulated and real world embeddings with increasing $N$ and dimensionality respectively. A window size of $N=100$ samples (2 seconds) was identified as providing the best trade-off between these factors.

\subsection*{Policy adaptation}\label{subsec:Method-Policy-Adaptation}
We adopt a similar approach to our previous work \cite{Cutting:SimToRealAdaptation} to adapt a pre-trained policy to observations synthesised from unlabelled target domain data. In this procedure, an ``expert'' policy $\pi_{e}$ is initially trained in a simulation environment with a physically-informed cutting model, as introduced in our previous work \cite{Cutting:LearningRoboticMilling}, with model parameters from Table \ref{tab:Method-Cutting-Model-Params}. The expert was trained initially for 32000 episodes using the proximal policy optimisation (PPO) algorithm with domain randomisation of material properties. A translation function $f: \mathbb{R}^{N\times N_{S}} \to \mathbb{R}^{N\times N_{A}}$ is applied to each state window:
\begin{equation}
x^{(g,i)*} = f(x^{(c,i)})
\end{equation}
and translated states paired with the corresponding expert action on $x^{(c)}$ to generate a labelled dataset
\begin{equation}
    \mathcal{D}=\{(x^{(g,i)},\pi_{e}(x^{(c,i)}))\}.
\end{equation}
We subsequently train a target domain policy $\pi_{g}$, initialised as $\pi_{g}=\pi_{e}$ on $\mathcal{D}$ using behavioural cloning. We note this procedure can be extended to alternative imitation learning algorithms (such as DAgger) provided $f$ can be inferred during generation of source windows $x^{(c,i)}$. Under the assumption that the environment satisfies the Markov property, the policy learning process is unaffected by the windowing procedure. As the full trajectories do not need to be reconstructed, limitations of other methods such as requirement for blending or enforcing temporal consistency are inapplicable to this work \cite{NeuralPolicyStyleTransferRL}. Furthermore, as each trajectory is decomposed into $T-N+1$ windows, the windowing approach has the effect of significantly augmenting the training data.

\subsection*{Style transfer}\label{subsec:Method-Style-Transfer}
\begin{figure}[!t]
    \centering
    \begin{subfigure}[t]{0.34\linewidth}
        \centering\raisebox{10mm}{\includegraphics[width=\linewidth]{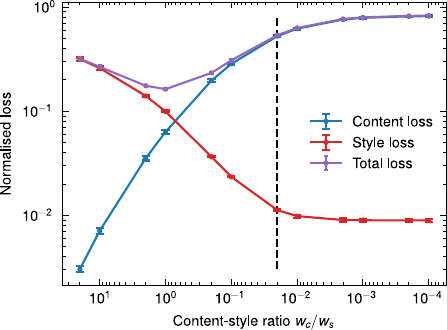}}
        \caption{}
    \end{subfigure}
    \begin{subfigure}[t]{0.65\linewidth}
        \centering\includegraphics[width=\linewidth]{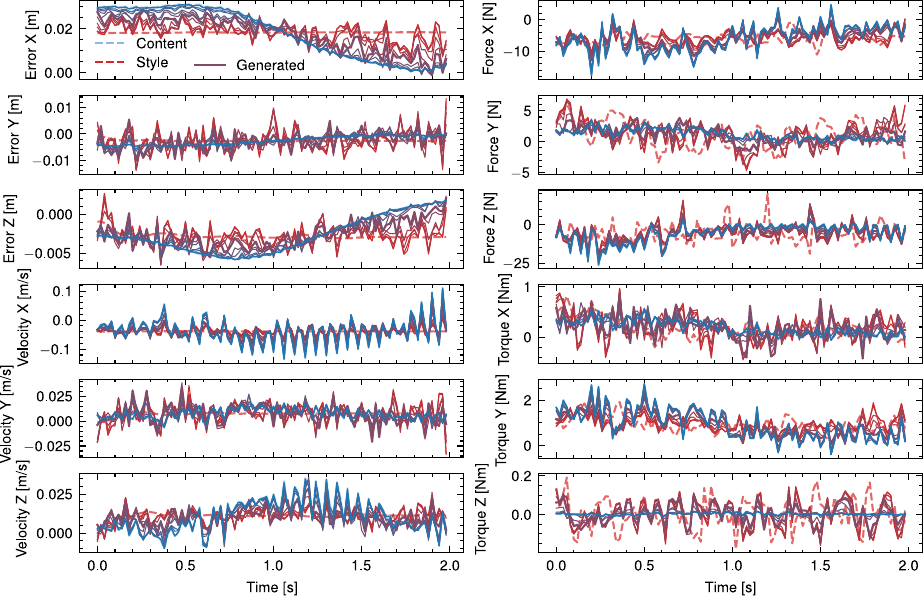}
        \caption{}
    \end{subfigure}
    \caption{Effect of content-style weight ratio $w_{c}/w_{l}$ on normalised (unweighted) content-style loss, averaged over 5 content-style batches (batch size 256), with chosen value $w_{c}/w_{s}=0.02$ indicated (dotted line). Decreasing ratio results in diminishing returns on style while diverging substantially from the original content windows. Increasing ratio tends towards identity (generated windows correspond to unaltered simulated windows).}
    \label{fig:Method-ST-Weights}
\end{figure}

In this work, we consider neural style transfer\cite{NeuralStyleTransferOriginal} as a translation function wherein $x^{(g)*}$ arise from solving the style transfer optimisation problem:
\begin{equation}
    x^{(g,i)*} = \arg\min_{x^{(g,i)}} \left( w_c L_c(x, x^{(c,i)}) + w_s L_s(x, x^{(s,j)}) \right)\label{eq:Method-Style-Transfer}
\end{equation}
where $w_{c}$ and $w_{s}$ are the content and style weights, respectively and $L_c$, $L_s$ are content and style loss contributions respectively. The content loss is defined as:
\begin{equation}
    L_{c} = \sum_{l} \sum_{i,j} \frac{1}{2N_{l}} \left(F^{(c,l)}_{ij}-F^{(g,l)}_{ij}\right)^2
\end{equation}
where $F^{(c,l)}$, $F^{(g,l)}$ are the feature outputs of layer $l$ for the content and generated output respectively. The style loss similarly is expressed as
\begin{equation}
    L_{s} = \sum_{l} \sum_{i,j} \frac{1}{4 N^{2}_{l} M^{2}_{l}} \left(G^{(s,l)}_{ij} - G^{(g,l)}_{ij}\right)^2
\end{equation}
where $\mat{G}^{(s,l)}$ is the style Gram matrix of layer $l$ outputs $F^{(s,l)}$
\begin{equation}
    \mat{G}^{(s,l)} = F^{(s,l)}F^{\tpose(s,l)}
\end{equation}
and similarly for $\mat{G}^{(c,l)}$. The generated windows were initialised as
\begin{equation}
    x^{(g,i)} = x^{(c,i)}
\end{equation}
and \eqref{eq:Method-Style-Transfer} optimised by gradient descent using the Adam optimiser. The relative content-style weighting $w_{c}/w{s}$ was tuned manually through a grid-search procedure. Figure \ref{fig:Method-ST-Weights} shows the effect of content-style weighting on their relative loss contributions. At low values of $w_{c}/w{s}$, the total loss is dominated by increasing content reconstruction error; the generated windows diverge substantially from the original windows with marginal effect on style reconstruction. Hence, $w_{c}/w{s}$ was reduced until diminishing returns on the (unweighted) style reconstruction loss was observed. We report relevant optimisation parameters in Table \ref{tab:Method-Hyperparameters}.

\begin{table}[t]
\centering
\begin{minipage}{0.48\textwidth}
    \centering
    \captionof{table}{Selected hyperparameters for encoder network and style transfer framework.}
    \begin{tabular}{ll}
    \hline
         Parameter & Value  \\
    \hline
         Encoder learning rate & $1\times10^{-3}$ \\
         Encoder output channels & $\left[128, 256, 512\right]$ \\
         Encoder kernel size & 3 \\
         Encoder batch size & 128 \\
         Encoder kernel stride & 2 \\
         Window size & 100 \\
         Latent dimensions & 130 \\
         Content-style ratio $w_{c}/w_{s}$ & 0.02 \\
         Style transfer learning rate & 0.01 \\
         Style transfer iterations & 1000 \\
         Content layer indices (Fig.~\ref{fig:Method-VAE-Architecture}) & [x] \\
         Style layer indices (Fig.~\ref{fig:Method-VAE-Architecture}) & [2, 5, 7] \\
    \hline
    \end{tabular}
    \label{tab:Method-Hyperparameters}
\end{minipage}
\hfill
\begin{minipage}{0.48\textwidth}
    \centering
    \includegraphics[width=\linewidth]{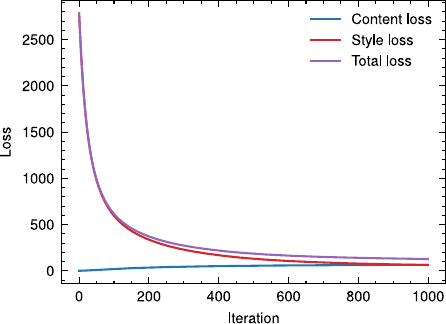}
    \captionof{figure}{Convergence plot for style transfer optimisation with parameters from Table~\ref{tab:Method-Hyperparameters} for a batch of 256 content-style pairings.}
    \label{fig:Method-ST-Convergence}
\end{minipage}
\end{table}

\begin{figure}[t]
\centering
\begin{minipage}[b]{0.48\textwidth}
    \centering
    \captionof{table}{Table of model parameters for cutting simulation (source domain)}
    \begin{tabular}{ll}
        \toprule
        \textbf{Parameter} & \textbf{Value} \\
        \midrule
        Pitch angle [rad] & 0.126 \\
        Helix angle [rad] & 0.0 \\
        Radius [m]     & 0.025 \\
        Cutter width [m] & 0.0005 \\
        Cutting elements (flutes) & 50 \\
        Spindle speed [rpm] & 1000 \\
        Material cutting\\-mechanistic constant ($K_c$) [N/mm$^{2}$] & \textit{variable} \\
        Material edge\\-mechanistic constant ($K_e$) [N/mm] & \textit{variable} \\
        \bottomrule
    \end{tabular}
    \label{tab:Method-Cutting-Model-Params}
\end{minipage}
\hspace{0.04\textwidth}
\begin{minipage}[t]{0.47\textwidth}
    \centering
    \includegraphics[width=\linewidth]{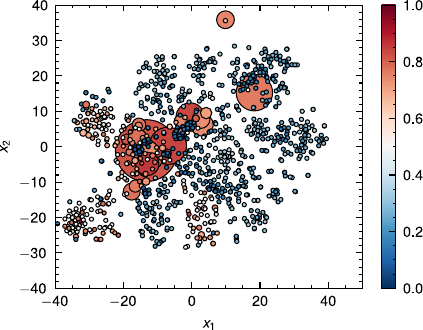}
    \captionof{figure}{t-SNE embedding diagram of content-style pairings. The points are coloured according to their class (simulation, blue / real world, red) with intensity according to the cosine similarity of their closest match, diverging from 0.5. The area of each real world embedding point is directly proportional to the number of times the corresponding window was matched.}
    \label{fig:Method-Autoencoder-VAE-Embeddings}
\end{minipage}
\end{figure}

A compelling advantage of encoder or classifier-based approaches is that they operate on unpaired cross-domain datasets. To improve the realism of generated trajectories, we employ a pairing mechanism that takes advantage of the unsupervised feature representations learned from the source domain data to generate weakly paired content and style windows. An intuitive analogue would be matching images with similar composition and subjects, reminiscent of the weak paring mechanism in \cite{InstaceStyleTransfer6DPose}. In the first stage, the real world dataset is encoded in entirety by the encoder network to generate a dataset of embeddings. In the second stage, the simulated content window(s) are encoded and a content-style pairing matrix is constructed by the pairwise cosine similarity between $x^{(c,i)}$, $x^{(s,j)}$ representations as
\begin{equation}
    S_{ij} = \frac{z_{i}\cdot z_{j}}{||z_{i}||\cdot||z_{j}||}
\end{equation}
In the last stage, the closest match real embedding is paired with the simulated embedding. For each row $i$, the index of the most similar pairing was obtained by:
\begin{equation}
    j^*i = \arg\max{j} , S_{ij}
\end{equation}

For windows where the pairing diverged substantially from the content, the optimisation process introduced mean shifts into the observations, as well as introducing artefacts from the encoding process. Following the intuition of \cite{ControllingPerceptualStyleTransfer}, qualitatively, we observed that pre-aligning the means of the content and style windows resulted in higher quality generated outputs. Figure \ref{fig:Method-Autoencoder-VAE-Embeddings} shows a representation of the content-style pairings generated by the pairing procedure. The data show the formation of distinct clusters according to simulated and real world trajectories. Unsurprisingly, the real world embeddings with the most matches were found predominantly at the intersections of the clusters. This parasitic behaviour is reminiscent of the mode-collapse phenomenon in generative-adversarial networks. Nonetheless, around 50\% of real world points were matched at least once, with matched windows dispersed throughout the latents, indicating good coverage of the real world dataset.

For adaptation, 50 episodic trajectories were collected in source domain with the expert policy, which formed the content dataset. For this work, the style dataset consisted of 148 off-policy trajectories collected from the real world. We note this is not a hard requirement; dataset size is motivated primarily by avoiding breakdown of the pairing and style transfer mechanism where content and style windows diverge substantially.

\subsection*{Experimental setup}\label{subsec:Method-Experimental}

As with our previous work, experimental validation was carried out on a KUKA LBR \textit{iiwa} R820 14kg collaborative robot equipped with a wrist-mounted motorised slitting saw tool. The \textit{iiwa} was connected via the Fast Research Interface (FRI) to a Robot Operating System (ROS) workstation with a communication frequency of 500Hz. The workstation consisted of an Intel i7-8086K CPU, NVIDIA GTX 1080 Ti GPU with 11GB VRAM, and 32GB RAM. The robot was equipped with a motorised slitting saw tool; whereas geometric parameters of the tool reflect the training parameters in Table \ref{tab:Method-Cutting-Model-Params}, the number of teeth was doubled to introduce further cross-domain mismatch.

The cutting task was represented as a single conventional milling pass over an material with variable geometry, following a nominal trajectory defined at the material surface. As proof of principle, the reference path was defined manually with respect to the surface for all case studies. During the cutting task, the policy provides as output a translational stiffness, incremental offset to the depth of cut (DoC), and the feed rate, relative to the planned (nominal) trajectory. The nominal feed rate was chosen as 0.75 m/min. The controller damping gain $\mat{K}_{d}$ was adjusted independently according to the stiffness to provide a damping ratio of 1.0 (i.e. critically damped). Trajectory tracking was achieved according to the operational space control law
\begin{equation}
	\vec{\Gamma} = \mat{J}^{\tpose}\left[\hat{\Lambda}(\vec{q})\left(\mat{K}_{d}(t)\dot{\vec{e}} + \mat{K}_{p}(t)\vec{e}\right) + \hat{\vec{\mu}}(\vec{q},\dot{\vec{q}})+\hat{\vec{\rho}}(\vec{q})\right]\label{eq:Method-OSC-Control-Law}
\end{equation}
where $\vec{\Gamma}$ are the commanded joint torques, $\mat{J}$ the robot Jacobian, and $\hat{\Lambda}$, $\hat{\vec{\mu}}$, $\hat{\vec{\rho}}$ are the estimated operational space inertia matrix, Coriolis \& centrifugal forces, and gravitational forces respectively.

During the cutting task, the process force was monitored via an FT-AXIA 80 force-torque sensor, mounted at the robot wrist. However, our method in principle is applicable to different types of sensors, such as those built in to the \textit{iiwa}, provided real world examples collected with such sensors. Prior to each trial, the force sensor was biased at the start of the trajectory. Force sensor gravity compensation was achieved via the following correction:
\begin{equation}
\vecframe{F}{W}_{ext}=\rotation{EE}{W}\vecframe{F}{EE}+mg\left(\unitvec{z}-\rotation{EE}{W}\rotation{W,0}{EE}\unitvec{z}\right)\label{eq:Method-Force-Sensor-Grav-Compensation}
\end{equation}
where $m$ is the tool mass, $g$ is the gravitational acceleration, $\vecframe{F}{W}_{ext}$ is the measured external force in the world frame $W$, $\unitvec{z}$ is the z-axis basis vector of $\refframe{W}$, and $\rotation{W}{EE}$, $\rotation{W,0}{EE}$ are the world to end-effector (EE) rotations at the current end-effector pose, and bias pose, respectively.

\begin{figure}
    \centering
    \includegraphics[width=0.9\textwidth]{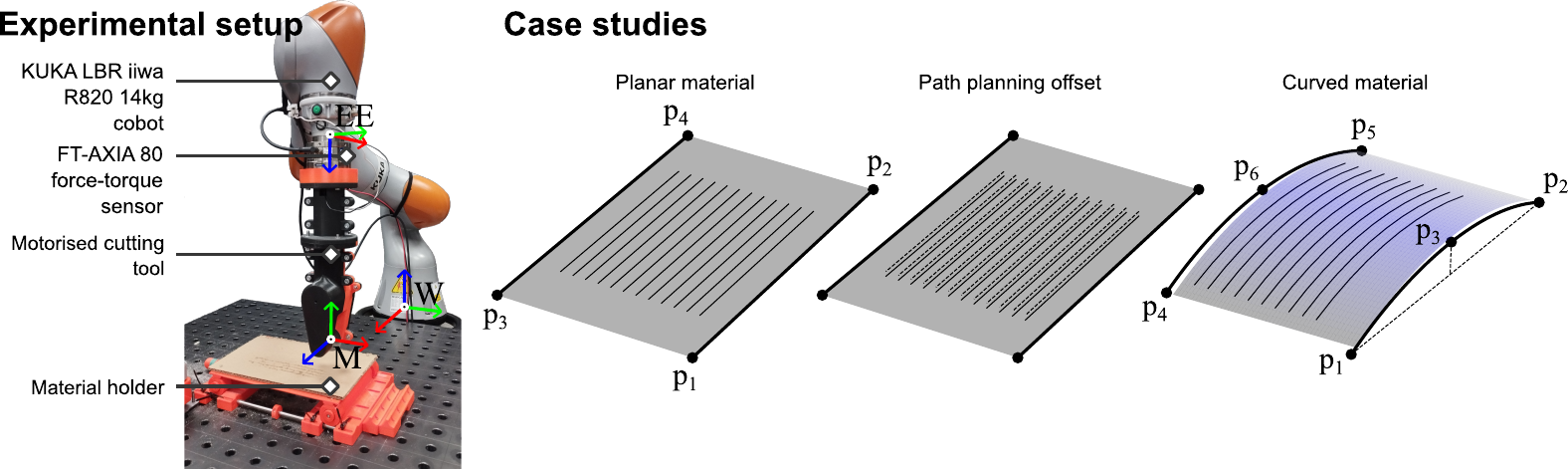}
    \caption{Overview of the experimental setup used for real world cutting experiments.}
    \label{fig:Method-Experimental-Setup}
\end{figure}

\section*{Results}

In this section, we evaluate the proposed method in comparison to the unadapted expert policy and state-of-the-art methods based on the previously established experimental setup. To demonstrate the performance of each method on a range of materials, cutting trials were carried out on polyurethane foam, cardboard, corrugated plastic, mica and aluminium. We further establish 3 separate case studies on each material to evaluate the policy performance under different path planning conditions. We evaluate each method by task completion time, average path deviation, average tool load, material removed volume (MRV), and similarity of the adopted action trajectories to the source domain expert actions, averaged over 5 trials per material for each strategy, and aggregated over all materials. To mitigate effects of drift (e.g. tool wear, temperature, calibration errors), trials for each strategy were interleaved.

\subsection*{Comparison methods}
For the subsequent real world experiments, we adopt the following terminology to denote comparison methods: `Expert' refers to the unadapted source simulation expert policy, as transferred directly to the real world task. `BC', or standalone behavioural cloning, represents our previous work, in which the simulation is augmented with a Gaussian process (GP) regression model trained on aligned demonstrations from 14 preliminary experiments on aluminium and mica. `CVAE' represents a conditional variational autoencoder using the same real world dataset as adopted for style transfer. Note in this instance, the encoder itself is trained on the entire dataset of both real world and simulation data, conditioned on a one-hot domain label (simulation or real world). Simulated data are encoded as with the style transfer approach, however, at decoding time, the one-hot class label is swapped to generate a synthetic window of the desired class. `CycleGAN' is also introduced as a comparison method. In this instance, the surrogate real world dataset is synthesised by the sim-to-real generator network. With all methods, the generator / encoder architecture was chosen equivalent to Table \ref{tab:Method-Hyperparameters}. For CycleGAN, a smaller discriminator network, with output channels $[64, 128, 256]$ was used due to mitigate the well-known `vanishing gradient' problem during GAN training. All other hyperparameters were chosen to be equivalent to the CycleGAN study. All methods were employed with behavioural cloning as per the self-supervision procedure introduced in this work. Additionally, as a benchmark, we include a ``baseline'' strategy in which the process parameters are held constant at the nominal feed rate (0.75m/min) and depth of cut of 1 mm, applied to all materials.

\subsection*{Planar material case study}
\begin{figure}[ht]
    \centering
    \begin{subfigure}[t]{\textwidth}
        \centering\includegraphics[width=\textwidth]{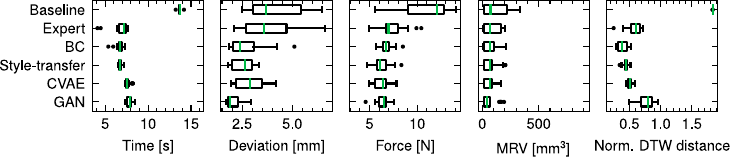}
        \caption{Flat}
        \label{fig:Results-Boxplot-ST-Strats-Summary-Flat}
    \end{subfigure}\\
    \begin{subfigure}[t]{\textwidth}
        \centering\includegraphics[width=\textwidth]{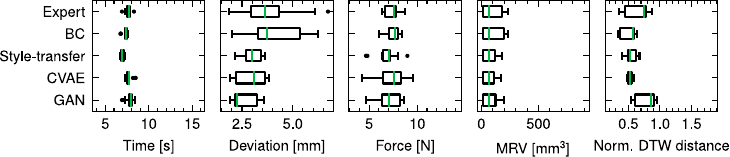}
        \caption{DoC offset}
        \label{fig:Results-Boxplot-ST-Strats-Summary-Doc1mm}
    \end{subfigure}\\
    \begin{subfigure}[t]{\textwidth}
        \centering\includegraphics[width=\textwidth]{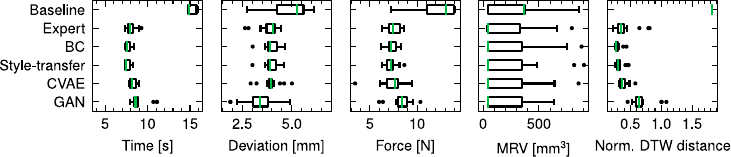}
        \caption{Curved}
        \label{fig:Results-Boxplot-ST-Strats-Summary-Curved}
    \end{subfigure}
    \caption{Boxplot summary of performance metrics for the style transfer trained policy and comparison methods, aggregated over all materials. Metrics include task completion time, average path deviation, average load force, average (normalised) dynamic time warping (DTW) distance between each strategy and the simulation expert policy (lower better), and material removed volume (MRV, higher better).}
    \label{fig:Results-Boxplot-ST-Strats-Summary}
\end{figure}

Each strategy was initially tested on a planar material, with the reference path calibrated at the material surface. For the calibration procedure, the surface was modelled as a warped plane interpolated between 4 corner points obtained via guarded move with a force threshold of 1N, with the exception of foam, where contact was confirmed visually. The performance of each strategy for the planar case study is outlined in Figure \ref{fig:Results-Boxplot-ST-Strats-Summary-Flat}. To aid interpretation, the significance of the difference in metrics was tested via one-way ANOVA. The normality and homoscedasticity assumptions of ANOVA were tested via the Shapiro-Wilk and Levene methods respectively. A significance level of $\alpha=0.05$ was used for all tests. Metrics that did not satisfy the assumptions were transformed via Box-Cox transform:
\begin{equation}
    y=\begin{cases}
    x^{\lambda}-1 & \mathrm{if}\,\lambda\neq0\\
    \log(x) & \mathrm{otherwise}
    \end{cases}
\end{equation}
where $\lambda$ is chosen to maximise the log-likelihood of the transformed data under a normality assumption. In the case of completion time and average force, the assumptions of ANOVA were satisfied (Shapiro $p=0.361$, $p=0.355$; Levene $p=0.0689$, $p=0.0983$, respectively). Average path deviation and MRV did not satisfy the normality assumption after transformation, and in this case the Kruskal-Wallis test was adopted without transformation. For both task completion time and average force, one-way ANOVA revealed significant effects of strategy on performance ($F=61.1$, $p=1.14\times10^{-27}$; $F=6.74$, $p=6.52\times10^{-5}$ respectively) between strategy and these performance metrics.

To examine the effect of individual strategy on the performance metrics, the Tukey Honestly Significant Difference (HSD) was used for ANOVA, and the Dunn post-hoc test for Kruskal-Wallis. No significant difference in task completion times was found between style transfer and BC, whereas the former outperformed all other methods. Style transfer had the largest effect relative to GAN ($-1.00$ s) and the smallest relative to the Expert ($-0.329$ s). For path deviation, style transfer significantly differed from the Expert ($-1.50$ mm, $p=0.000196$) and GAN ($0.451$ mm, $p=0.005074$) strategies, however, results were inconclusive for BC ($p=0.560$) and CVAE ($p=0.109$). Style transfer was further found to significantly outperform the Expert and BC strategies in minimising average force ($-1.273$ N, $p=0.0001$; $-0.651$ N, $p=0.0352$), however, no significant difference was found between style transfer and the CVAE and GAN strategies ($p=0.867$, $p=0.611$). The choice of strategy was found to have no conclusive effect on MRV (Kruskal $H=2.87$, $p=0.578$). This result appears surprising in light of the differing action selection apparent for each strategy, particularly in DoC.

To examine the effect of the adaptation methods on the agent actions, the actions taken during each trial were compared with 50 simulated experiments (i.e. source domain) carried out with the source domain expert, and the similarity of action trajectories evaluated by normalised dynamic time warping (DTW) distance. The strategies that adopt actions that are more broadly similar to the source domain expert will score lower on this metric than those that deviate substantially from the expert behaviour. The expert policy itself was included in this comparison since it is being applied to the \emph{target} domain. We report effect sizes as Hedges' $g$. Clear differences between the strategies were indicated (Kruskal $H=1930$, $p=0.0$), with style transfer yielding large improvements relative to the Expert $g=0.875$ and GAN $g=2.18$, a moderate improvement for CVAE $g=0.575$ and a small reduction in performance relative to BC $g=-0.370$. Post-hoc testing indicated a high significance level in these effects ($p\leq3.65\times10^{-17}$) for all comparisons.

\begin{figure}[p]
    \centering
    \includegraphics[width=0.75\textwidth]{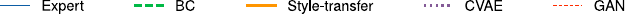}\\
    {\sf\hspace{0.04\textwidth}Flat\hspace{0.27\textwidth}DoC offset\hspace{0.25\textwidth}Curved}\\
    \vspace{0.1cm}
    \begin{subfigure}{0.32\textwidth}
        \centering\includegraphics[width=\textwidth]{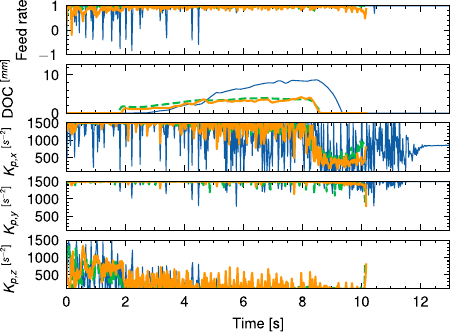}
        \caption{Foam - BC, style transfer}
        \label{fig:Results-Action-Comparisons-Flat-Foam-1}
    \end{subfigure}
    \begin{subfigure}{0.32\textwidth}
        \centering\includegraphics[width=\textwidth]{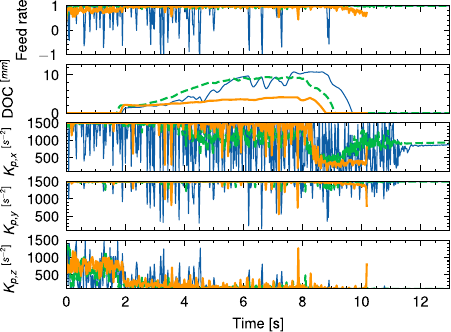}
        \caption{Foam - BC, style transfer}
        \label{fig:Results-Action-Comparisons-Doc1mm-Foam-1}
    \end{subfigure}
    \begin{subfigure}{0.32\textwidth}
        \centering\includegraphics[width=\textwidth]{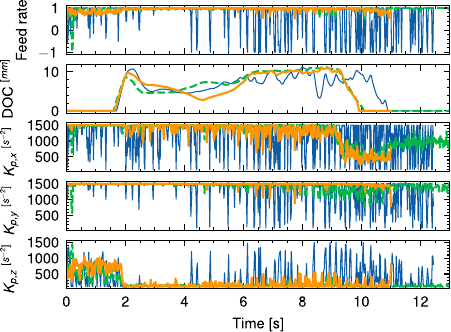}
        \caption{Foam - BC, style transfer}
        \label{fig:Results-Action-Comparisons-Curved-Foam-1}
    \end{subfigure}\\
    
    \begin{subfigure}{0.32\textwidth}
        \centering\includegraphics[width=\textwidth]{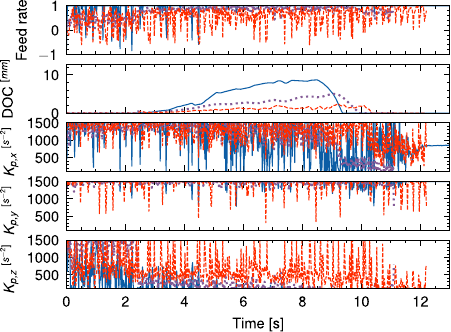}
        \caption{Foam - CVAE, GAN}
        \label{fig:Results-Action-Comparisons-Flat-Foam-2}
    \end{subfigure}
    \begin{subfigure}{0.32\textwidth}
        \centering\includegraphics[width=\textwidth]{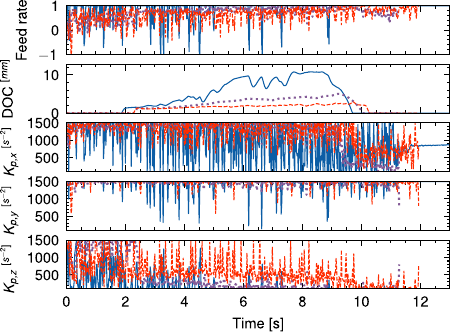}
        \caption{Foam - CVAE, GAN}
        \label{fig:Results-Action-Comparisons-Doc1mm-Foam-2}
    \end{subfigure}
    \begin{subfigure}{0.32\textwidth}
        \centering\includegraphics[width=\textwidth]{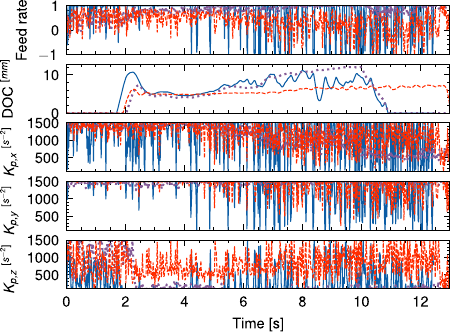}
        \caption{Foam - CVAE, GAN}
        \label{fig:Results-Action-Comparisons-Curved-Foam-2}
    \end{subfigure}\\

    \begin{subfigure}{0.32\textwidth}
        \centering\includegraphics[width=\textwidth]{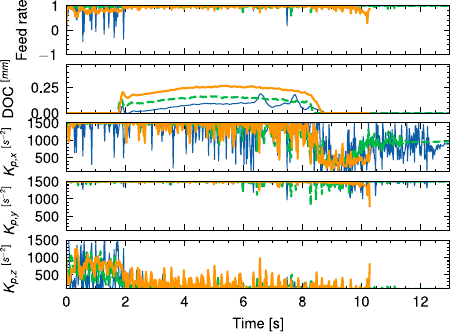}
        \caption{Mica - BC, style transfer}
        \label{fig:Results-Action-Comparisons-Flat-Mica-1}
    \end{subfigure}
    \begin{subfigure}{0.32\textwidth}
        \centering\includegraphics[width=\textwidth]{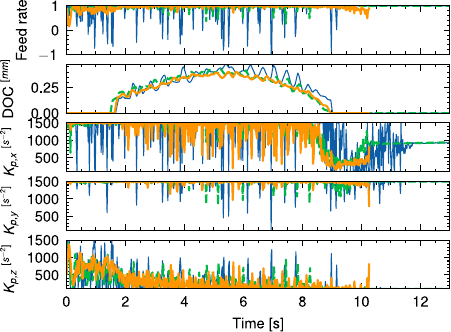}
        \caption{Mica - BC, style transfer}
        \label{fig:Results-Action-Comparisons-Doc1mm-Mica-1}
    \end{subfigure}
    \begin{subfigure}{0.32\textwidth}
        \centering\includegraphics[width=\textwidth]{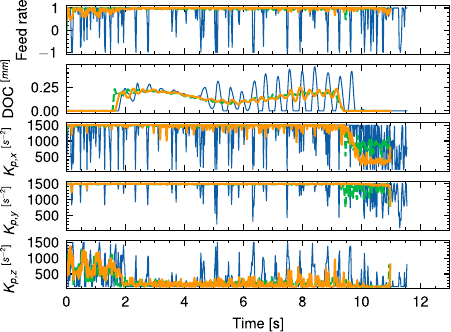}
        \caption{Mica - BC, style transfer}
        \label{fig:Results-Action-Comparisons-Curved-Mica-1}
    \end{subfigure}\\

    \begin{subfigure}{0.32\textwidth}
        \centering\includegraphics[width=\textwidth]{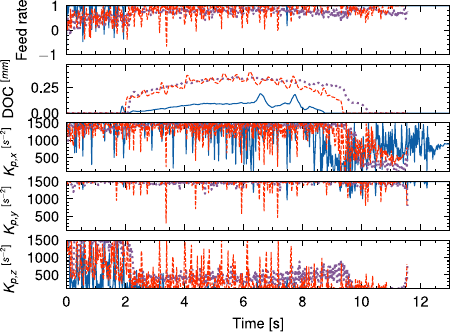}
        \caption{Mica - CVAE, GAN}
        \label{fig:Results-Action-Comparisons-Flat-Mica-2}
    \end{subfigure}
    \begin{subfigure}{0.32\textwidth}
        \centering\includegraphics[width=\textwidth]{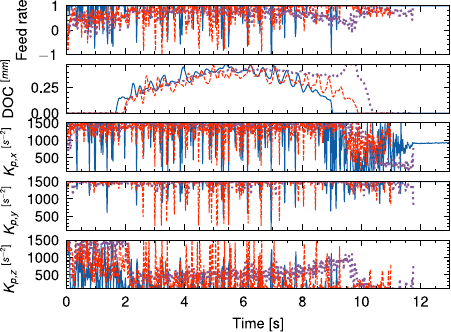}
        \caption{Mica - CVAE, GAN}
        \label{fig:Results-Action-Comparisons-Doc1mm-Mica-2}
    \end{subfigure}
    \begin{subfigure}{0.32\textwidth}
        \centering\includegraphics[width=\textwidth]{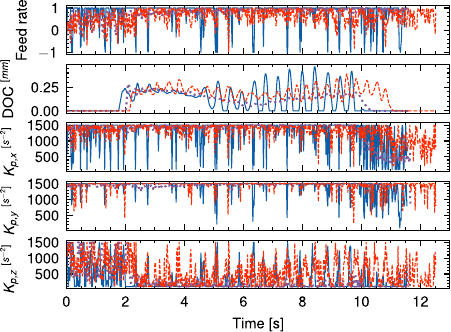}
        \caption{Mica - CVAE, GAN}
        \label{fig:Results-Action-Comparisons-Curved-Mica-2}
    \end{subfigure}
    \caption{Comparison of agent actions for foam and mica for planar, DoC offset and curved case studies respectively. Actions include the relative to nominal feed rate adjustment, with 0 corresponding to no change and 1 to double the nominal feed rate, depth of cut (DoC), and controller stiffness in transverse, feed direction and normal directions respectively ($K_{p,x}$, $K_{p,y}$, $K_{p,z}$. Units of $K_{p}$ are chosen consistent with \eqref{eq:Method-OSC-Control-Law}).}
    \label{fig:Results-Action-Comparisons}
\end{figure}

To examine the behaviour of each strategy in more detail and enable qualitative comparisons between each strategy, the action trajectories adopted by each policy during an example trial on foam and mica are presented in Figure \ref{fig:Results-Action-Comparisons}. From Figure \ref{fig:Results-Action-Comparisons-Flat-Foam-1}, \ref{fig:Results-Action-Comparisons-Flat-Foam-2}, \ref{fig:Results-Action-Comparisons-Flat-Mica-1}, \ref{fig:Results-Action-Comparisons-Flat-Mica-2}, the action trajectories were broadly similar between BC and style transfer across both materials. Style transfer adopts a more correct behaviour of reducing the feed rate prior to engagement with the material, as compared with BC. Conversely, the GAN policy diverges substantially from the expert behaviour which corroborates the DTW metric results. All adapted policies adopted a more consistent DoC throughout both trials than the unadapted expert policy. Differences between the policy behaviour on each material were mainly evident in the DoC behaviour, transverse stiffness ($K_{x}$) and, to a lesser extent, the normal stiffness ($K_{z}$).

\subsection*{Robustness to path planning offset}

To examine the robustness of the method to path planning errors - for example, due to errors in calibration, surface position estimation or noise, we further examine the performance of all strategies with a path planning offset of 1mm, inset into the ground truth material surface. The performance of each strategy was evaluated over 3 trials per strategy, per material.

Similarly to the planar case, strategy significantly impacted task completion times (ANOVA $F=17.0$, $p=9.94\times10^{-10}$); style transfer again differed significantly from all strategies except BC (Tukey HSD $p=0.0694$), and improvements over other strategies being similar to the planar case study, albeit more consistent across strategies (effect size range $-0.658$-$-0.850$ s). Whereas path deviation was also influenced by strategy (ANOVA $F=4.61$, $p=0.00235$), post-hoc testing indicated only GAN differed significantly from the BC ($p=0.0047$) and Expert ($p=0.0125$) strategies. Although group means were more concentrated than in the planar case, path deviation was notably more consistent across trials for style transfer, CVAE, and GAN, implying these strategies were better able to tolerate the path planning offset and maintain stable path tracking across materials. MRV was again unaffected by strategy (Kruskal-Wallis $H=2.61$, $p=0.624$), and contrasting the planar case study, no significant differences were observed in average tool load (ANOVA $F=1.06$, $p=0.382$). Similarly to the planar case study, there was a clear separation between the strategies in terms of similarity to expert actions (Kruskal $H=688$, $p=9.33\times10^{-148}$). Post-hoc testing indicated style transfer was distinct from the comparison methods, with the least significant result being with CVAE ($p=0.0314$), small negative effects for BC $g=-0.321$ and CVAE $g=-0.151$ and positive effects relative to Expert $g=0.682$ and GAN $g=1.31$ strategies.

Figure \ref{fig:Results-Action-Comparisons-Doc1mm-Foam-1}, \ref{fig:Results-Action-Comparisons-Doc1mm-Foam-2}, \ref{fig:Results-Action-Comparisons-Doc1mm-Mica-1}, \ref{fig:Results-Action-Comparisons-Doc1mm-Mica-2} shows the agent actions for the offset case study. All strategies exhibited a more sporadic DoC behaviour than the planar case study, with style transfer exhibiting the most consistent DoC behaviour across both materials, and matching more closely to the planar case study behaviour, supporting observations regarding the consistency of the path deviation. All strategies exhibited a more aggressive variation in stiffness relative to the planar case study, indicating a compensatory response to the offset cutting depth.

\subsection*{Non-planar surfaces}

We further showcase the performance of each strategy when both material and surface geometry are altered to varying degrees of curvature. Consistent with the planar case study, the reference path with respect to the surface was assumed already known; however, we note that numerous path-planning methods have been proposed in the context of milling, including the case where surface geometry is unknown \cite{Cutting:RoboticMilling3DPointCloud}. For this case study, we assume the material is a thin plate under pure bending, with the surface modelled as a section of a truncated oblique cone -- in other words, an interpolation between two circular arcs. The arc parameters for each endpoint were derived from a 3-point estimation obtained similarly to the planar case study. Curvatures ranged between 2.36 m$^{-1}$ and 4.04 m$^{-1}$ across materials. Cardboard was excluded from the set of materials since the maximum curvature generated during preliminary experiments did not meaningfully differ from the previous case studies.

\begin{figure}[!t]
	\centering
    \begin{subfigure}[t]{0.4\textwidth}
        \centering\includegraphics[width=\textwidth]{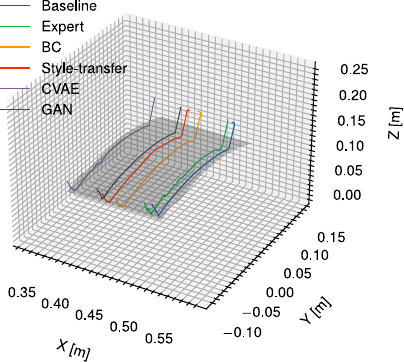}
        \caption{Polyurethane foam}
        \label{fig:Results-Curved-3D-Foam}
    \end{subfigure}
    \begin{subfigure}[t]{0.4\textwidth}
        \centering\includegraphics[width=\textwidth]{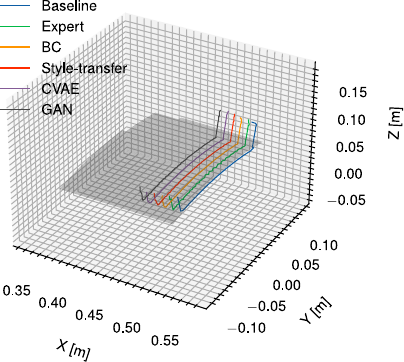}
        \caption{Corrugated plastic}
        \label{fig:Results-Curved-3D-Plastic}
    \end{subfigure}\\\smallskip
    \begin{subfigure}[t]{0.4\textwidth}
        \centering\includegraphics[width=\textwidth]{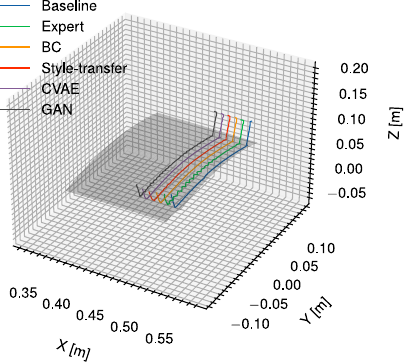}
        \caption{Mica}
        \label{fig:Results-Curved-3D-Mica}
    \end{subfigure}
    \begin{subfigure}[t]{0.4\textwidth}
        \centering\includegraphics[width=\textwidth]{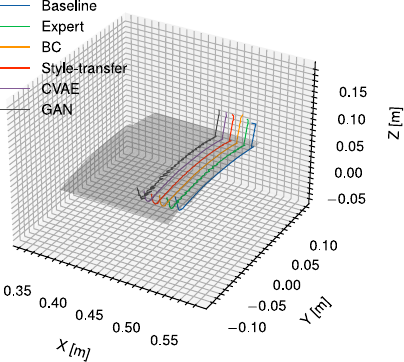}
        \caption{Aluminium}
        \label{fig:Results-Curved-3D-Aluminium}
    \end{subfigure}
	\caption{3D plot of TCP paths adopted by each strategy with respect to the material surface - qualitative defects are shown in the ``expert'' and ``GAN'' strategies, which exhibit transverse path deviations on the stiffer materials.}
	\label{fig:Results-Curved-3D-Paths}
\end{figure}

As with the prior case studies, strategy had a significant effect on completion time (Kruskal $H=38.5$, $p=8.44\times10^{-8}$) and in post-hoc testing, style transfer outperformed all strategies except BC ($p=0.529$). The effect of style transfer largely reflected the planar case study, with $-1.00$ s relative to GAN, and $-0.413$ s relative to the Expert. Differences in path deviation were inconclusive compared to the planar case study, (Kruskal $H=9.77$, $p=0.0445$) with the most significant result from post-hoc testing arising between GAN and style transfer ($p=0.0589$); however, differences in average force were more pronounced (ANOVA $F=7.71$, $p=0.000025$), with style transfer significantly outperforming GAN ($-1.25$ N, $p=0.0001$) but not the other strategies. Corroborating the previous case studies, MRV did not significantly differ between strategies (Kruskal $H=2.15$, $p=0.708$). Furthermore, action similarity again revealed clear separation between strategies (Kruskal $H=1390$, $p=2.64\times10^{-300}$), with style transfer exhibiting the largest deviation from GAN ($g=2.14$) and significant differences from all others (BC $g=-0.264$, CVAE $g=0.503$, Expert $g=0.487$). 

The agent actions, as shown in Figure \ref{fig:Results-Action-Comparisons-Curved-Foam-1}, \ref{fig:Results-Action-Comparisons-Curved-Foam-2}, \ref{fig:Results-Action-Comparisons-Curved-Mica-1}, \ref{fig:Results-Action-Comparisons-Curved-Mica-2}, show similar behaviours to the offset case study, with differences in DoC behaviour becoming more pronounced, particularly for the expert policy. CycleGAN adopted a highly sporadic action profile in feed rate and stiffness, particularly for the foam trials. A hypothesis for this behaviour is that the curved material presents a more challenging case for the agent and the much lower cutting forces limit information available to the agent to make decisions. Therefore, the actions resemble those at the beginning of the planar trials in which the agent is in free space and has no information about the contact state or tool engagement. Style transfer and BC both exhibited less consistent DoC behaviour than the planar case studies on foam, however, produced smoother action trajectories that were strongly correlated to the engagement state - for example, contact initiation was well-demarcated for both strategies.

Figure \ref{fig:Results-Curved-3D-Paths} shows a representative example of the 3D TCP positions adopted by each strategy for a single cutting trial. The TCP trajectories adopted exhibited clear defects for the expert and GAN trials, which were evident across both low and high stiffness materials. On the low stiffness materials, such as in Figure \ref{fig:Results-Curved-3D-Foam} these were evident as low-frequency irregularities, resembling a random walk, whereas for the high stiffness materials, this was exhibited as a higher frequency ``wobble'', which were unrelated to known phenomena such as chattering. These defects were suppressed or entirely absent during the BC, CVAE and style transfer trials, with these methods yielding similar qualitative improvements across all materials.

\section*{Discussion}

For the cutting task, the proposed method was evaluated based on task completion times, average path deviation, tool load (average force), material removed volume, behavioural similarity to expert action trajectories in source domain, and qualitatively by the action trajectories, ability to maintain consistent cutting conditions (e.g. depth of cut), as well as TCP trajectories. Relative to the comparison methods -- consisting of the unadapted source domain expert policy (Expert), our previous work (BC), conditional variational autoencoder (CVAE) and CycleGAN (generative adversarial network) -- the proposed method based on style transfer consistently achieved significant reductions in task completion time across all case studies. Compared to BC and CVAE, style transfer showed comparative performance but did not uniformly surpass them across all metrics.

The reduced influence of strategy in the offset path case study is consistent with the constraint imposed by insetting the path into the material, which limits the ability of the agent to regulate the true DoC. It also implies a common limitation of these methods in modelling out-of-distribution task conditions, wherein offsetting the reference path and nominal feed rate introduces concept shift in the optimal actions across domains in addition to covariate shift in the observations. Although path deviation was more consistent across style transfer, CVAE and GAN strategies than for BC and the expert policy, overall improvements were primarily inconclusive. It is plausible that the inconclusive effects may be attributable to the reduced number of samples for the offset case study.

Qualitatively, the style transfer trained policy demonstrated improved behavioural stability relative to the model-free approaches, with smoother action trajectories and more consistent control of depth-of-cut and stiffness, which was robust to perturbations in surface geometry and cutting path, and corroborated by higher action similarity to the source domain expert relative to all strategies except BC. The irregular path deviations observed in the TCP trajectories were attributable to the largely sporadic action trajectories of the expert policy, and, to a lesser extent, the GAN strategy. For the stiffer materials, deviations in the path are caused by contact instabilities resulting from interaction between the policy stiffness and the environment stiffness. These behaviours were largely absent with the BC, CVAE and style transfer strategies.

We hypothesise that the poorer performance of CycleGAN-based domain adaptation arises from its limited capacity to preserve task-relevant structure in the translated observations, which has been documented in related work \cite{TimeSeriesTranslationCNN}. While CycleGAN has been effective in visual domains where semantic content remains invariant under style changes---e.g. image-to-image translation, its application to time-series control tasks may disrupt temporal dependencies or distort dynamics-critical features, leading to degraded policy performance.

\section*{Conclusion}
An example-based approach for sim-to-real transfer in robotic control was proposed based on the principle of neural style transfer. Empirical results on a robotic cutting task demonstrate that the proposed method achieves comparable or superior performance to our previous work, conditional variational autoencoders, and CycleGAN-based time series translation across diverse materials and geometric scenarios, while substantially relaxing the assumptions of our previous example-based work. The proposed method is sample-efficient, demonstrated with 148 off-policy real world trajectories versus 32000 for initial policy training, and avoids the need for training domain discriminator, generator or corrective models, a crucial limitation of previously proposed adaptation methods.

We note the limitation that this work does not explicitly address differing cross-domain target (action) distributions or compatibility of generated trajectories with robot kinematic and dynamic constraints. We posit such constraints could be formulated as part of the optimisation process wherein physical feasibility losses are jointly optimised with style and content losses, and represents a possible extension of this work. Additionally, the quality of generated trajectories and pairings is expected to deteriorate with low coverage of real-world examples, weak content-style match similarity, or parasitic matching where a small subset of real trajectories dominate the pairing.

\section*{Data availability}
The datasets generated during and/or analysed during the current study are available in the Figshare repository, DOI 10.6084/m9.figshare.28983659.

\section*{Funding Declaration}
This work was supported by the UK Research and Innovation (UKRI) project “Research and Development of a Highly Automated and Safe Streamlined Process for Increase Lithium-ion Battery Repurposing and Recycling” (REBELION) under Grant 101104241. 

\section*{Acknowledgements}
The authors would further like to acknowledge Abdelaziz Wasfy Shaarawy, Carl Meggs and Christopher Gell respectively for assistance with experimental validation, design of material holder and cutter tool for experiments herein.

\section*{Author contributions}
Conceptualisation - A.R. and J.H.; data curation - J.H.; formal analysis - J.H.; funding acquisition - A.R. and R.S.; investigation - J.H.; methodology - J.H. and A.R.; project administration - A.R. and R.S.; software - J.H.; resources - J.H.,  A.R. and R.S.; supervision - A.R. and R.S.; validation - J.H. and A.R.; visualisation - J.H.; writing (original draft) - J.H.; writing (review \& editing) - J.H. and A.R. and R.S.

\section*{Competing interests}
The authors declare no competing interests.

\bibliography{literature}

\end{document}